\documentclass{article}

\usepackage[utf8]{inputenc} 
\usepackage[T1]{fontenc}    
\usepackage{hyperref}       
\usepackage{url}            
\usepackage{booktabs}       
\usepackage{amsfonts}       
\usepackage{nicefrac}       
\usepackage{microtype}      
\usepackage{xcolor}         
\usepackage{graphicx}
\usepackage{amsmath}
\usepackage{titlesec}
\usepackage{natbib}         

\titleformat*{\section}{\Large\bfseries}
\titleformat*{\subsection}{\large\bfseries}

\usepackage[margin=1in]{geometry}

\title{In Context Learning with Vision Transformers: Case Study}

\author{
    Antony Zhao \quad 
    Alex Proshkin \quad 
    Fergal Hennessy \quad 
    Francesco Crivelli \\
    \small UC Berkeley \\
    \small \texttt{ayzhao7761@berkeley.edu, pro@berkeley.edu,} \\
    \small \texttt{fergalhennessy@berkeley.edu, crivelli@berkeley.edu}
}

\begin{document}

\maketitle

\begin{abstract}
  Large transformer models have been shown to be capable of performing in-context learning. By using examples in a prompt as well as a query, they are capable of performing tasks such as few-shot, one-shot, or zero-shot learning to output the corresponding answer to this query. One area of interest to us is that these transformer models have been shown to be capable of learning the general class of certain functions, such as linear functions and small 2-layer neural networks, on random data (Garg et al, 2023). We aim to extend this to the image space to analyze their capability to in-context learn more complex functions on the image space, such as convolutional neural networks and other methods.
\end{abstract}

\section{Introduction}

In-context learning is a capability of large transformer models, such as some large language models like ChatGPT. It allows the model to learn from the samples in the prompt to predict some output without having to train on these samples directly. \cite{garg2023} showed that this performance can be comparable to an almost optimal solver, such as comparing its ability to learn the general class of linear functions against the performance of a least squares algorithm. We aim to extend this to the image space to further investigate and increase our understanding of in-context learning.

Recently, Vision Transformers have been shown to be noteworthy for their small inductive bias. They have exceeded conventional Convolutional Neural Networks on many benchmarks, but they often struggle to learn meaningful representations on small datasets.

In-Context Learning has been shown to allow transformers to infer latent concepts and patterns at test time, significantly improving accuracy. Notably, it has been shown by Min, Lyu, Holtzman et al. (\href{https://arxiv.org/abs/2202.12837}{ArXiv}) that providing a strong pattern of input-output context significantly improves test results roughly the same amount, regardless of whether the context mappings match ground truth. In this paper, we apply the techniques from in-context learning to vision transformers (ViT) and assess their performance, in comparison to convolutional networks trained on the same data, allowing us to analyze the potential to use these models on smaller datasets by utilizing ICL.

\section{Related Work}

\textbf{CNN vs ViT:}

Bai et al. (2021) conduct a rigorous evaluation of the robustness of Vision Transformers (ViTs) compared to Convolutional Neural Networks (CNNs) under standardized training conditions. Their findings indicate that CNNs can achieve adversarial robustness on par with ViTs when trained using similar methodologies. Additionally, while ViTs exhibit superior generalization to out-of-distribution data, this advantage is primarily attributed to their self-attention mechanisms rather than pretraining on large-scale datasets. This study underscores the importance of fair experimental setups in assessing model robustness and provides insights into the architectural factors influencing performance. \href{https://arxiv.org/abs/2111.05464}{ArXiv}

\textbf{ViT Transfer Learning vs Training from Scratch:}

Steiner et al. (2021) perform a comprehensive empirical study of Vision Transformers (ViT), isolating how training-data scale, data-augmentation/regularization ("AugReg"), model size and compute budget interact/ By training more than 50 000 ViT variants on ImageNet-21k and JFT-300M under a unified protocol, they show that carefully tuned AugReg combined with additional compute can recover the accuracy gains normally obtained by a 10 × larger pre-training dataset, and that these gains transfer to downstream tasks after fine-tuning. The study also quantifies compute–performance trade-offs and releases all checkpoints, providing practical guidelines for data- and compute-efficient ViT training. \href{https://arxiv.org/pdf/2106.10270}{ArXiv}

\textbf{Are Transformers More Robust Than CNNs?}

Bai et al. (2021) conduct the first rigorously controlled comparison of model robustness between Vision Transformers and CNNs by matching model capacity (DeiT-S vs. ResNet-50) and unifying all training hyperparameters, optimizers, and data-augmentation schedules. They show that when CNNs inherit the strong augmentation and AdamW/weight-decay "Transformer" recipe, their adversarial robustness (PGD, AutoAttack, texture-patch attacks) becomes indistinguishable from that of Transformers, overturning earlier claims based on mismatched experimental setups. In contrast, Transformers still exhibit superior out-of-distribution generalization on ImageNet-C, ImageNet-A and Stylized-ImageNet; ablations attribute this benefit to the self-attention architecture itself rather than to large-scale external pre-training or other training heuristics. The study also introduces an augmentation-warm-up curriculum that stabilizes adversarial training for ViTs and releases complete code and checkpoints, establishing a reproducible baseline for future robustness research. 
\href{https://arxiv.org/pdf/2111.05464}{ArXiv}

\textbf{Few-Shot Vision Transformers (ICL)}

Dong et al. (2022) revisit few-shot visual classification with Vision Transformers and identify that, unlike CNNs, vanilla ViTs struggle to learn reliable patch-token dependencies when only a handful of labelled examples are available. They introduce Self-promoted sUpervisioN (SUN), which first self-pretrains a ViT on the target few-shot dataset and then uses the resulting teacher to create location-specific similarity labels that act as additional patch-level supervision during meta-training. Two refinements—background-patch filtration and spatial-consistent augmentation—further enhance the quality of the generated local labels. Evaluated with Meta-Baseline on mini-ImageNet, tiered-ImageNet and CIFAR-FS (5-way, 1-/5-shot), SUN lifts ViT accuracy above previous ViT baselines and, for the first time, surpasses strong CNN counterparts, demonstrating that appropriate token-wise guidance can compensate for Transformers' lack of inductive bias in low-data regimes. \href{https://arxiv.org/pdf/2203.07057}{ArXiv}

\section{Problem Statement}

 Extending the findings of \cite{garg2023}, our goal is to extend in-context learning performance gains to the image space. Using input-output pairs sampled from the dataset and sample functions, we generate a series of images and embeddings and pass them into our Transformer model as context for our final output. We investigate various models for our Transformer over different function classes, such as randomized convolutional and ViT models.

\section{Methods}

We first construct a transformer model that is capable of handling both images and embeddings. For the prompt, we pass the sequence of $(x_1, f(x_1), x_2, f(x_2), ...x_k,f(x_k)$, where $f(\cdot)$ is a sample function from some function class, such as linear or convolutions, to our model. This model is then trained to predict the $f(x_i)$ sequence, while masking the following values to ensure it only has the preceding context as information for each prediction. We process images by passing them through a ViT or CNN to obtain image embeddings. We then pass these image embeddings alongside the $f(x_i)$ embeddings produced by the sample function or model into a decoder-only GPT-2 style model and obtain the predictions for each of the $f(x_i)$. These two models, (the ViT or CNN, and the GPT-2 model) are trained alongside each other, and are randomly initialized at the start.

We sample $x_i$ from the set of CIFAR-10 images, which we downscale from a $32\times32$ image to a $8\times8$ image. This is also normalized by subtracting the mean and standard deviation across the entire dataset. We chose to use these downscaled images rather than random noise as there are inherent patterns in the image that could be interesting to study.

\begin{figure}[htbp]
    \centering
    \includegraphics[width=0.4\linewidth]{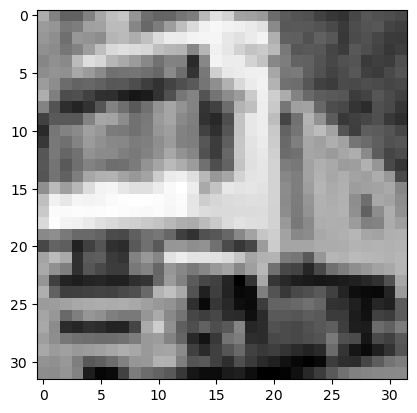}
    \includegraphics[width=0.4\linewidth]{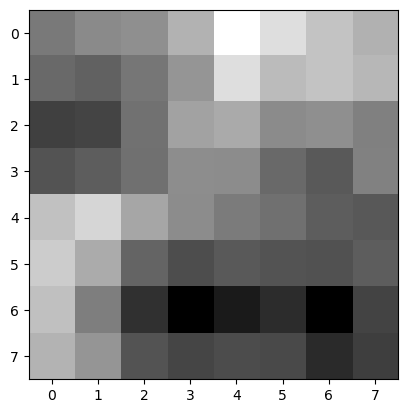}
    \caption{Example image from the CIFAR10 and the corresponding downscaled image.}
    \label{fig:cifar_example}
\end{figure}

We use a variety of function classes for our sample functions. These include randomized linear functions, randomized convolutional networks, and randomized ViT using sinusoidal positional embeddings.

Similar to \cite{garg2023}, we employ curriculum training by starting with a low-dimensional image by masking other values outside of a square $d\times d$ block to 0 and slowly increasing the dimensionality of the image as well as the number of samples used in the prompt. Generally, we increment $d$ by one each time and use $n=kd+1$ samples, where $k$ is some value such as $5$ or $8$.

\begin{figure}[htbp]
    \centering
    \includegraphics[width=0.5\linewidth]{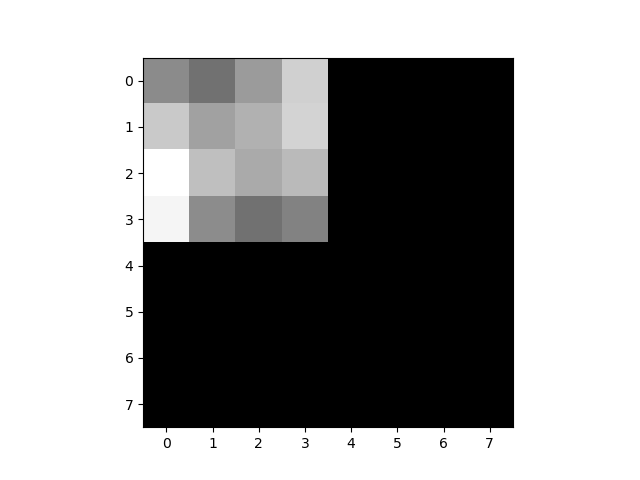}
    \caption{An example of the masking process, where we take one of the images and then zero out (which is shown as blacking out) anything outside of some $d\times d$ block.}
    \label{fig:masking_example}
\end{figure}

The model was then trained to predict each $f(x_i)$ using mean squared error loss, and we used masking to ensure $f(x_i)$ would only have access to $(x_1, f(x_1),...x_i)$ to make the prediction. We also weighted the loss so that it would focus more on the second half of $f(x_i)$, which improved performance on the final query prompt. This weighting mask can be modeled as:

\begin{equation}
    \left( \frac{2k}{n} \right)^2 \quad \text{for } k = 1, 2, \dots, n
\end{equation}

\section{Experiments}
\label{sec:experiments}

This section evaluates a decoder‑only transformer's ability to perform \emph{in‑context} regression on image data under four progressively harder settings.  Each experiment is tagged \texttt{E1}–\texttt{E4}.  Unless otherwise noted, curriculum training follows Section~4 and all models are evaluated with mean‑squared error (MSE).

\subsection{In‑Context Linear Function Approximation}

\subsubsection*{E1–CNN+GPT}
\label{exp:e1}
A CNN encoder is used to produce the image embeddings.  We draw $w\sim\mathcal N(0,I_{d^2})$ and set $f(x)=w^Tx$.  The prompt contains $n=5d{+}1$ example pairs; $d$ is slowly increased from 2 to 8 by one every 5000 update steps. Training was allowed to continue for a full 500k steps, but following experiments will halt training early as the loss plateaus much earlier.

\subsubsection*{E2 – ViT Encoder}
\label{exp:e2}
Identical to E1 except the image encoder is a 4‑layer, 2‑head ViT with a patch size of 4 and learned positional embeddings.  Convergence is reached after \textasciitilde300k steps.

\begin{figure}[htbp]
  \centering
  \includegraphics[width=0.9\linewidth]{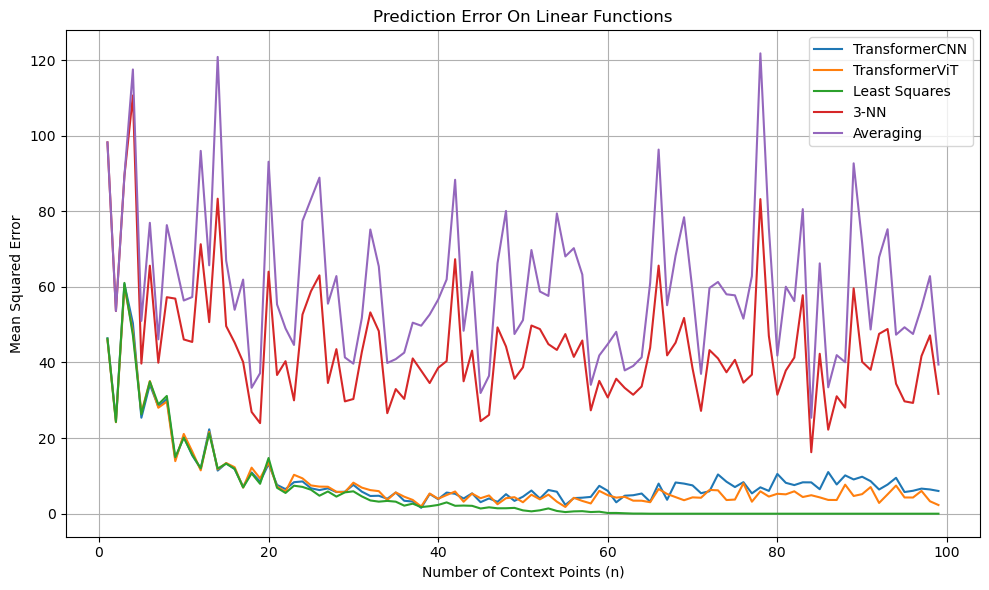}
  \caption{Linear regression results (E1 and E2).  Lower is better.}
  \label{fig:lin_results}
\end{figure}

\paragraph{Results.} Figure~\ref{fig:lin_results} compares E1 and E2 against least‑squares, 3‑nearest‑neighbors, and mean baselines.  Both transformer variants approach the pseudo‑inverse solution and outperform non‑parametric baselines.

\begin{figure}[htbp]
  \centering
  \includegraphics[width=0.9\linewidth]{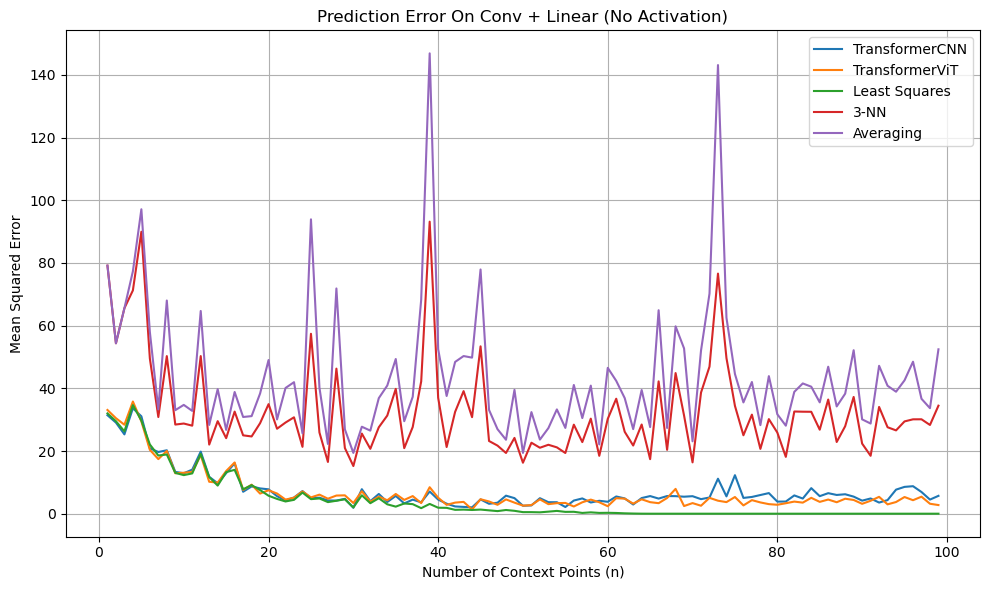}
  \caption{Out‑of‑distribution linear convolution results.}
  \label{fig:lin_ood}
\end{figure}

We additionally evaluate on a shifted distribution where $f(x)=w^T(k*x)$ with an orthogonally‑initialized $3\times3$ kernel $k$.  Performance remains comparable (Figure~\ref{fig:lin_ood}).

\subsection{In‑Context Non‑Linear Function Approximation}

\subsubsection*{E3 – Random 2‑Layer CNN Target}
\label{exp:e3}
Targets are generated by a frozen two‑layer CNN $f(x)=w^T\alpha(k*x)$ with $k\sim\mathcal N(0,I_{2\times2})$ and $w\sim\mathcal N(0,I_{16})$, and $\alpha$ is the ReLU activation. The transformer architecture and curriculum mirror E2.

\subsubsection*{E4 – Random ViT Target}
\label{exp:e4}
Here $f(\cdot)$ is a frozen 4‑layer ViT with sinusoidal positional encodings. Since we predicted that this task would be more complex, we increased the number of layers of the ViT inside our model to 12 (though this appeared to have been unnecessary). Because target activations are lower‑magnitude, training converges after 80k updates.

\begin{figure}[htbp]
  \centering
  \includegraphics[width=0.9\linewidth]{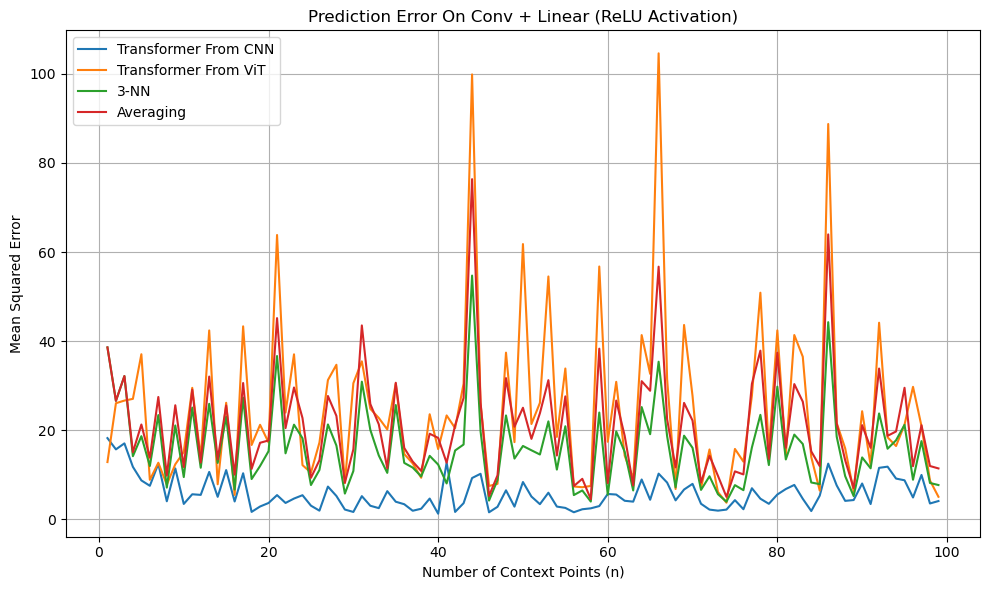}
  \caption{Prediction error on convolutional targets (E3).}
  \label{fig:cnn_target}
\end{figure}

\begin{figure}[htbp]
  \centering
  \includegraphics[width=0.9\linewidth]{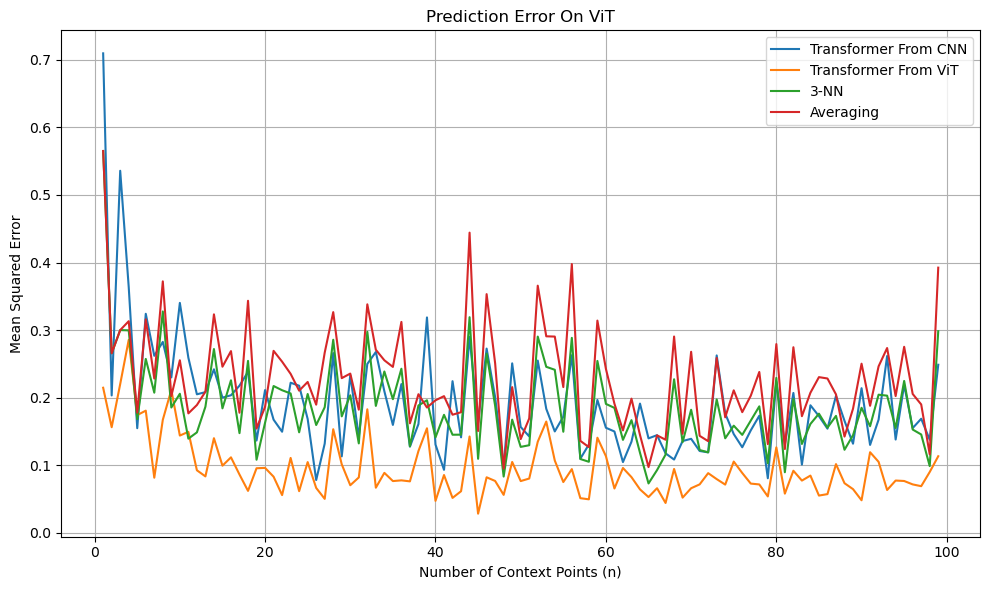}
  \caption{Prediction error on ViT targets (E4).}
  \label{fig:vit_target}
\end{figure}

\paragraph{Results.}  Figures~\ref{fig:cnn_target} and \ref{fig:vit_target} show that the in‑context learner matches or exceeds fresh models trained from scratch (MLP, CNN, ViT) on the same support examples, particularly when only a few context pairs are available.

\begin{figure}[htbp]
  \centering
  \includegraphics[width=0.9\linewidth]{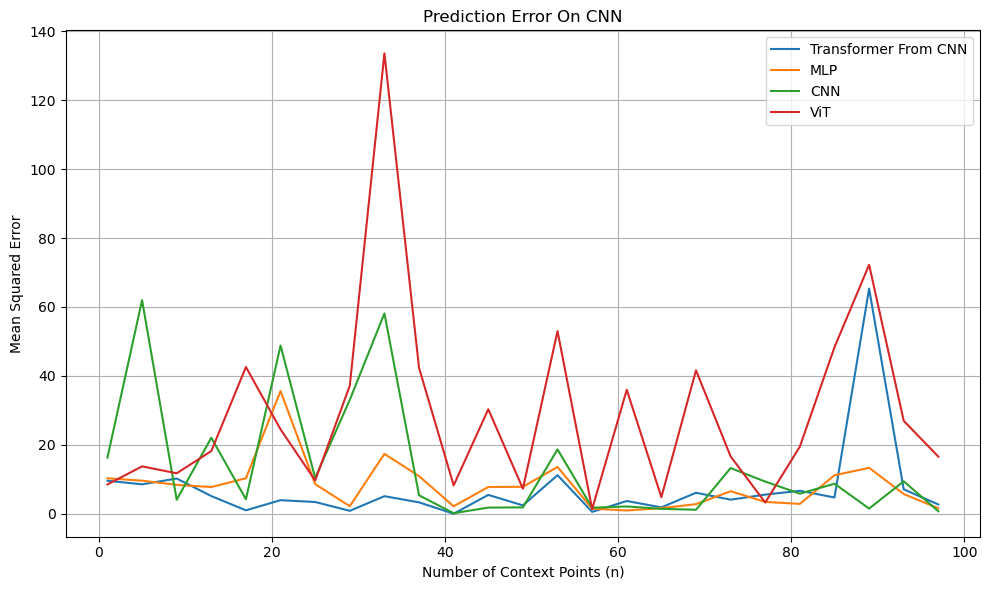}
  \caption{E3 vs. gradient‑descent‑trained baselines.}
  \label{fig:cnn_gd}
\end{figure}

\begin{figure}[htbp]
  \centering
  \includegraphics[width=0.9\linewidth]{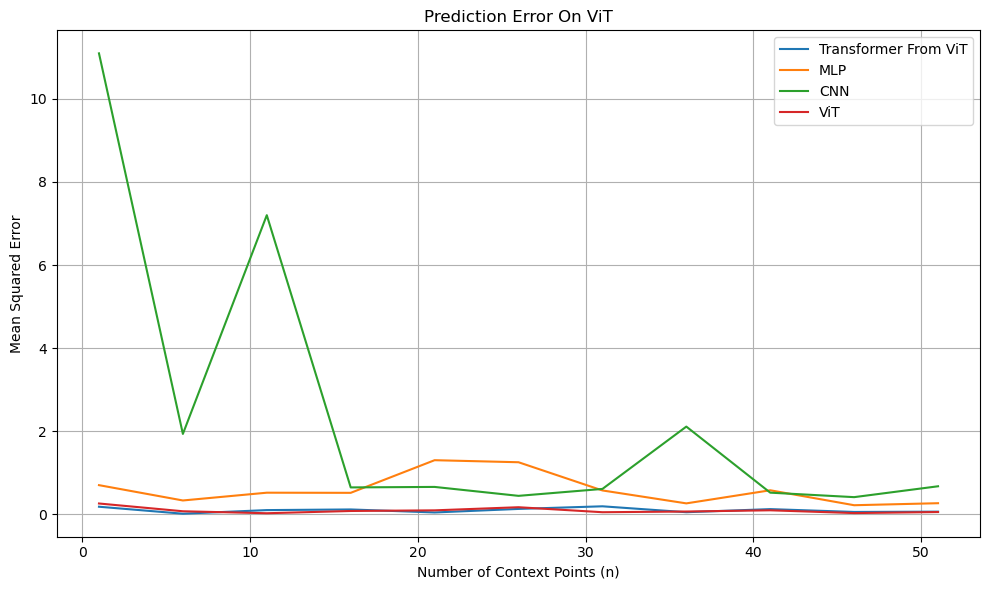}
  \caption{E4 vs. gradient‑descent‑trained baselines.}
  \label{fig:vit_gd}
\end{figure}

\paragraph{Comparison with Gradient‑Descent Training.}  Figures~\ref{fig:cnn_gd} and \ref{fig:vit_gd} benchmark against models that receive the same $n$ samples and are trained for 5000 steps with Adam (learning rate of $10^{-3}$).

\section{Code}

The code used to generate this document is available on \href{https://github.com/antony-zhao/182-final-project}{Github}

\section{Discussion}
Our results show that a decoder‐only transformer can in‐context learn simple convolutional mappings on small grayscale images, achieving low MSE with only a handful of examples. However, as input size grows, more context is needed and convergence slows, indicating sample‐complexity challenges. We only had the compute for testing relatively simple models on much smaller images; extending to deeper, multi‐channel models and larger or more natural images remains future work. It could also be interesting to investigate ICL on a set of trained models, which would be able to exploit the image features. Exploring larger transformer backbones, such as alternative fusion (e.g., cross‐attention), and optimized curricula may reduce the number of required examples and improve generalization.

\bibliographystyle{plain}
\bibliography{references}


\newpage

\section{Appendix}
\subsection{Hyperparameters}

\begin{table}[htbp]
\centering
\begin{tabular}{c c c} 
 \hline
 Embedding Size & \#Layers & \#Heads\\ [0.5ex] 
 \hline\hline
    256 & 12 & 8 \\
  \hline
\end{tabular}
\caption{GPT-2 Model}
\end{table}

\begin{table}[htbp]
\centering\begin{tabular}{c c c c c c} 
 \hline
 Experiment & \#Convolutional/ViT Layers & \#Channels/Heads & Kernel/Patch Size  & Learning Rate\\ [0.5ex] 
 \hline\hline
 E1 & 8 Conv Layers & 32 Channels & 3 Kernel & $10^{-4}$ \\ 
  \hline
  E2 & 4 ViT Layers & 8 Heads & 4 Patch & $10^{-4}$ \\ 
  \hline
  E3 & 4 ViT Layers & 8 Heads & 4 Patch & $10^{-4}$ \\ 
  \hline
  E4 & 12 ViT Layers & 8 Heads & 4 Patch & $10^{-5}$ \\ 
  \hline
\end{tabular}
\caption{Experiment Hyperparameters}
\end{table}

\subsection{Training Results}

\begin{figure}[htbp]
    \centering
    \includegraphics[width=0.4\linewidth]{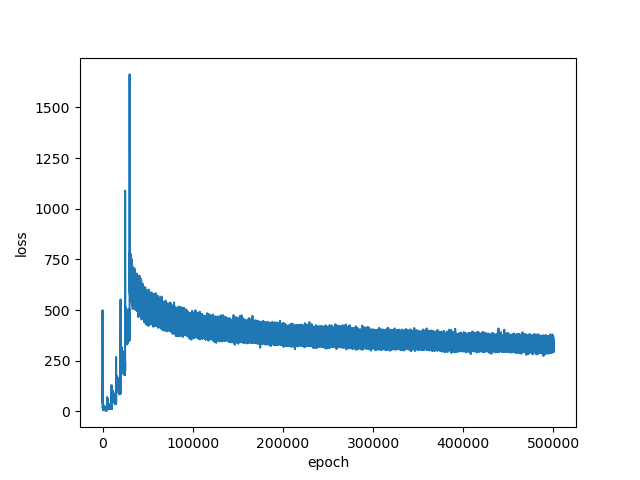}
    \includegraphics[width=0.4\linewidth]{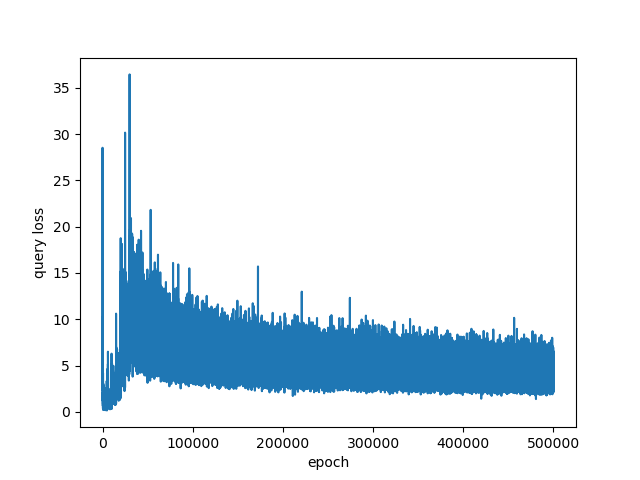}
    \caption{Loss curves from training experiment 1 (CNN + Transformer on linear functions) over 500k epochs.}
    \label{fig:exp1_loss}
\end{figure}

\begin{figure}[htbp]
    \centering
    \includegraphics[width=0.4\linewidth]{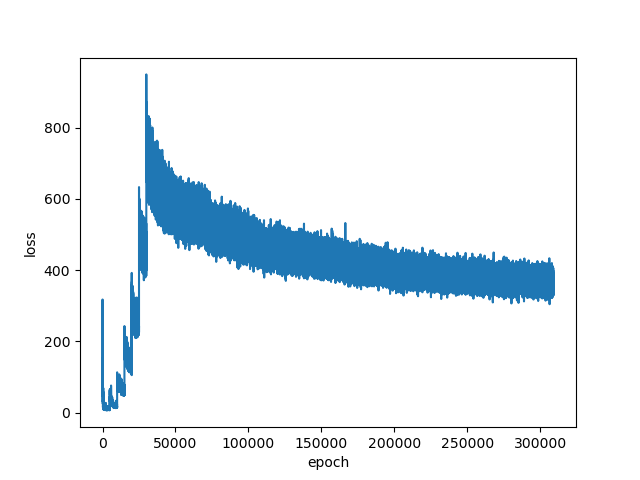}
    \includegraphics[width=0.4\linewidth]{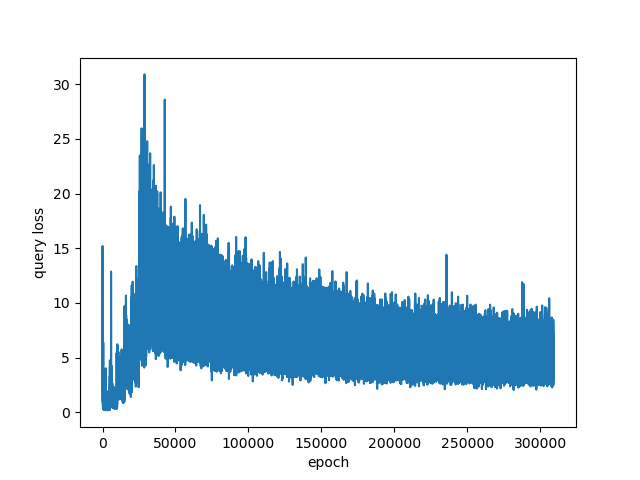}
    \caption{Loss curves from training experiment 2 (ViT + Transformer on linear functions) over 300k epochs.}
    \label{fig:exp2_loss}
\end{figure}

\begin{figure}[htbp]
    \centering
    \includegraphics[width=0.4\linewidth]{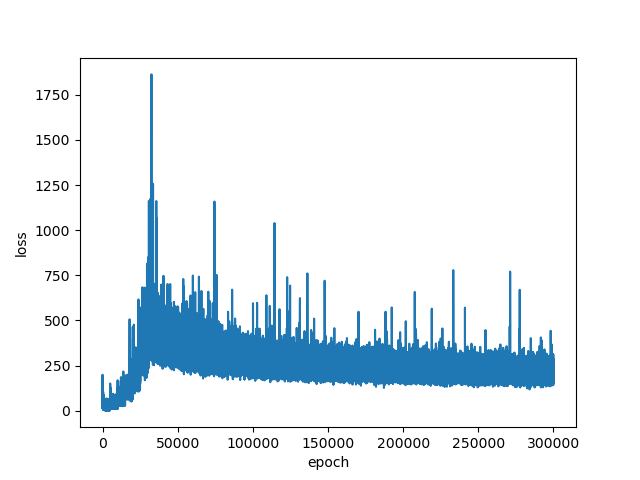}
    \includegraphics[width=0.4\linewidth]{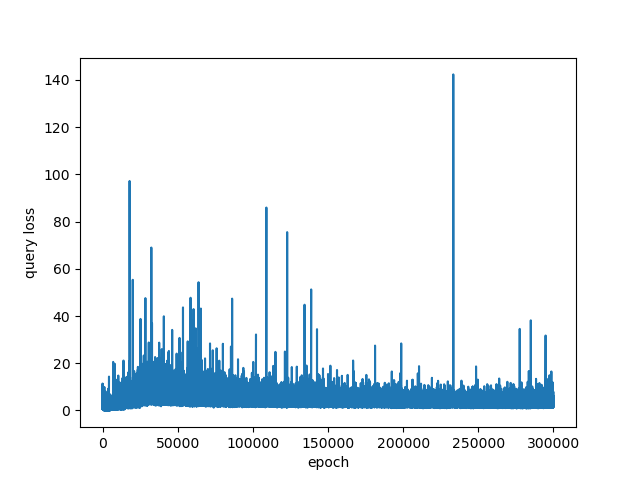}
    \caption{Loss curves from training experiment 3 (ViT + Transformer on convolutional networks) over 300k epochs.}
    \label{fig:exp3_loss}
\end{figure}

\begin{figure}[htbp]
    \centering
    \includegraphics[width=0.4\linewidth]{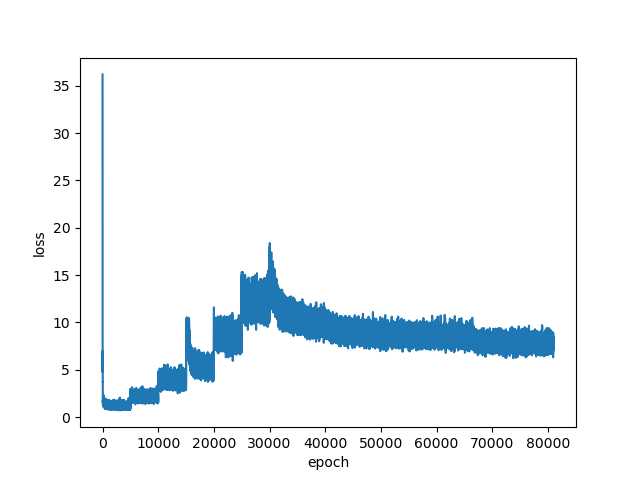}
    \includegraphics[width=0.4\linewidth]{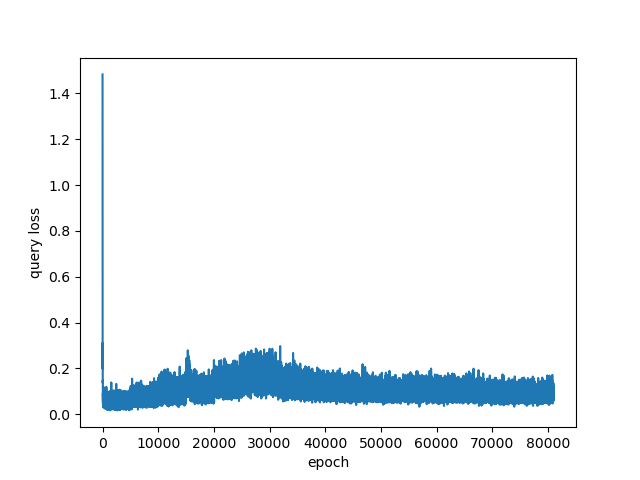}
    \caption{Loss curves from training experiment 4 (ViT + Transformer on ViTs) over 80k epochs.}
    \label{fig:exp4_loss}
\end{figure}
\end{document}